\documentclass[10pt, a4paper]{article}
\usepackage{lrec}
\usepackage{graphicx}
\usepackage{tabularx}
\usepackage{soul}

\usepackage{epstopdf}
\usepackage[latin1]{inputenc}

\usepackage{hyperref}
\usepackage{xstring}

\usepackage{booktabs}
\newcommand\blfootnote[1]{%
  \begingroup
  \renewcommand\thefootnote{}\footnote{#1}%
  \addtocounter{footnote}{-1}%
  \endgroup
}

\usepackage{color}

\title{CoVoST: A Diverse Multilingual Speech-To-Text Translation Corpus}

\name{Changhan Wang, Juan Pino, Anne Wu*, Jiatao Gu}

\address{Facebook AI \\
         \{changhan, juancarabina, annewu, jgu\}@fb.com\\}

\abstract{
Spoken language translation has recently witnessed a resurgence in popularity, thanks to the development of end-to-end models and the creation of new corpora, such as Augmented LibriSpeech~\cite{kocabiyikoglu2018augmenting} and MuST-C~\cite{di-gangi-etal-2019-must}. Existing datasets involve language pairs with English as a source language, involve very specific domains or are low resource. We introduce CoVoST, a multilingual speech-to-text translation corpus from 11 languages into English, diversified with over 11,000 speakers and over 60 accents. We describe the dataset creation methodology and provide empirical evidence of the quality of the data. We also provide initial benchmarks, including, to our knowledge, the first end-to-end many-to-one multilingual models for spoken language translation. CoVoST is released under CC0 license and free to use. We also provide additional evaluation data derived from Tatoeba under CC licenses. \\ \newline  }

\begin{document}

\maketitleabstract
\blfootnote{\textbf{*} Work done as part of the Facebook AI Residency.}

\section{Introduction}
End-to-end speech-to-text translation (ST) has attracted much attention recently~\cite{alex2016listen,duong-etal-2016-attentional,Weiss_2017,Bansal_2017,berard2018end} given its simplicity against cascading automatic speech recognition (ASR) and machine translation (MT) systems. The lack of labeled data, however, has become a major blocker for bridging the performance gaps between end-to-end models and cascading systems. Several corpora have been developed in recent years. \newcite{post2013improved}
introduced a 180-hour Spanish-English ST corpus by augmenting the transcripts of the Fisher and Callhome corpora with English translations. \newcite{di-gangi-etal-2019-must} created the largest ST corpus to date from TED talks but the language pairs involved are out of English only. \newcite{beilharz2019librivoxdeen} created a 110-hour German-English ST corpus from LibriVox audiobooks. \newcite{godard-etal-2018-low} created a Moboshi-French ST corpus as part of a rare language documentation effort. \newcite{woldeyohannis} provided an Amharic-English ST corpus in the tourism domain. \newcite{boito2019mass} created a multilingual ST corpus involving 8 languages from a multilingual speech corpus based on Bible readings~\cite{cmu_wilderness}. Previous work
either involves language pairs out of English, very specific domains, very low resource languages or a limited set of language pairs. This limits the scope of study, including the latest explorations on end-to-end multilingual ST~\cite{inaguma2019multilingual,gangi2019onetomany}. Our work is mostly similar and concurrent to \newcite{iranzosnchez2019europarlst} who created a multilingual ST corpus from the European Parliament proceedings. The corpus we introduce has larger speech durations and more translation tokens. It is diversified with multiple speakers per transcript/translation. Finally, we provide additional out-of-domain test sets.

In this paper, we introduce CoVoST, a multilingual ST corpus based on Common Voice~\cite{ardila2019common} for 11 languages into English, diversified with over 11,000 speakers and over 60 accents. It includes a total 708 hours of French (Fr), German (De), Dutch (Nl), Russian (Ru), Spanish (Es), Italian (It), Turkish (Tr), Persian (Fa), Swedish (Sv), Mongolian (Mn) and Chinese (Zh) speeches, with French and German ones having the largest durations among existing public corpora. We also collect an additional evaluation corpus from Tatoeba\footnote{https://tatoeba.org/eng/downloads} for French, German, Dutch, Russian and Spanish, resulting in a total of 9.3 hours of speech. Both corpora are created at the sentence level and do not require additional alignments or segmentation. Using the official Common Voice train-development-test split, we also provide baseline models, including, to our knowledge, the first end-to-end many-to-one multilingual ST models. CoVoST is released under CC0 license and free to use. The Tatoeba evaluation samples are also available under friendly CC licenses. All the data can be obtained at \url{https://github.com/facebookresearch/covost}.

\begin{table*}[t]
    \small
    \centering
    \begin{tabular}{cc|c|cc|cc|cc|cc|cc}
    & & & \multicolumn{2}{c|}{Sentences} & \multicolumn{2}{c|}{Speaker} & \multicolumn{2}{c|}{Tokens} & \multicolumn{2}{c|}{Average Length} & \multicolumn{2}{c}{Word Vocab} \\
    & & Hours & All & Unique & Count & Accents & Source & Target & Source & Target & Source & Target \\
    \toprule
    & Train & 87.1 & 78.9K & 27.5K & 436 & 9 & 787.7K & 800.8K & 10.0 & 10.1 & 29.7K & 25.3K  \\
    Fr & Dev & 38.3 & 34.1K & 10.4K & 1,001 & 17 & 336.0K & 339.0K & 9.8 & 9.9 & 14.6K & 12.8K  \\
    & Test & 46.3 & 39.2K & 10.4K & 2,884 & 24 & 391.6K & 392.0K & 10.0 & 10.0 & 14.9K & 13.2K  \\
    & TT & 1.6 & 4.5K & 4.5K & 3 & N/A & 25.6K & 24.4K & 5.7 & 5.4 & 3.4K & 2.2K  \\
    \midrule
    & Train & 71.0 & 60.3K & 8.5K & 1,109 & 7 & 549.5K & 605.5K & 9.1 & 10.0 & 16.3K & 11.8K \\
    De & Dev & 88.1 & 77.3K & 5.6K & 2,337 & 11 & 690.8K & 759.2K & 8.9 & 9.8 & 12.0K & 9.3K \\
    & Test & 168.3 & 145.8K & 5.6K & 4,781 & 13 & 1.31M & 1.43M & 9.0 & 9.8 & 12.3K & 9.5K \\
    & TT & 4.0 & 9.1K & 9.1K & 5 & N/A & 45.8K & 47.0K & 5.0 & 5.1 & 4.9K & 3.2K \\
    \midrule
    & Train & 4.4 & 4.3K & 1.9K & 35 & 2 & 39.9K & 41.5K & 9.4 & 9.8 & 9.2K & 7.7K \\
    Nl & Dev & 5.3 & 5.0K & 1.7K & 126 & 2 & 48.0K & 50.0K & 9.4 & 9.8 & 4.3K & 4.0K \\
    & Test & 8.2 & 7.7K & 1.7K & 461 & 3 & 73.6K & 76.5K & 9.5 & 9.9 & 4.3K & 3.9K \\
    & TT & 0.3 & 0.6K & 0.6K & 1 & N/A & 2.9K & 3.2K & 5.1 & 5.5 & 0.7K & 0.7K \\
    \midrule
    & Train & 10.2 & 7.1K & 2.1K & 6 & N/A & 75.2K & 91.2K & 10.6 & 12.8 & 7.4K & 4.8K \\
    Ru & Dev & 9.0 & 6.4K & 1.7K & 9 & N/A & 66.3K & 80.5K & 10.4 & 12.7 & 6.5K & 4.3K \\
    & Test & 8.2 & 5.8K & 1.7K & 61 & N/A & 59.6K & 72.3K & 10.3 & 12.5 & 6.2K & 4.1K \\
    & TT & 1.5 & 2.7K & 2.7K & 5 & N/A & 15.2K & 18.4K & 5.7 & 6.9 & 4.2K & 2.7K \\
    \midrule
    & Train & 20.9 & 18.3K & 6.9K & 319 & 11 & 162.8K & 177.3K & 8.9 & 9.7 & 5.6K & 4.5K \\
    Es & Dev & 3.2 & 2.7K & 2.6K & 89 & 10 & 24.5K & 26.6K & 9.0 & 9.8 & 5.2K & 4.2K \\
    & Test & 3.5 & 2.7K & 2.6K & 457 & 10 & 24.2K & 26.4K & 8.8 & 9.6 & 5.2K & 4.1K \\
    & TT & 1.9 & 2.8K & 2.8K & 2 & N/A & 22.2K & 23.6K & 7.8 & 8.3 & 4.2K & 3.3K \\
    \midrule
    & Train & 13.4 & 10.0K & 6.4K & 28 & 1 & 116.7K & 127.8K & 11.8 & 12.9 & 12.8K & 9.9K \\
    It & Dev & 10.6 & 8.3K & 4.6K & 93 & 1 & 92.8K & 103.1K & 11.2 & 12.4 & 10.6K & 8.1K \\
    & Test & 12.8 & 8.9K & 4.6K & 577 & 1 & 100.8K & 110.3K & 11.4 & 12.5 & 10.4K & 8.1K \\
    \midrule
    & Train & 2.6 & 2.5K & 1.8K & 14 & 1 & 18.5K & 24.6K & 7.3 & 9.7 & 4.7K & 3.4K \\
    Tr & Dev & 3.0 & 2.9K & 1.6K & 58 & 1 & 21.0K & 28.1K & 7.2 & 9.6 & 4.3K & 3.1K \\
    & Test & 3.8 & 3.4K & 1.6K & 323 & 1 & 24.7K & 33.2K & 7.2 & 9.7 & 4.2K & 3.1K \\
    \midrule
    & Train & 19.9 & 16.2K & 2.4K & 352 & N/A & 133.8K & 164.9K & 8.3 & 10.2 & 5.5K & 3.9K \\
    Fa & Dev & 22.8 & 18.4K & 2.1K & 677 & N/A & 150.8K & 185.0K & 8.2 & 10.1 & 5.1K & 3.7K \\
    & Test & 23.9 & 19.1K & 2.1K & 1,210 & N/A & 157.9K & 193.5K & 8.3 & 10.2 & 5.1K & 3.7K \\
    \midrule
    & Train & 1.2 & 1.6K & 1.6K & 2 & N/A & 10.9K & 12.2K & 6.8 & 7.6 & 2.3K & 2.0K \\
    Sv & Dev & 1.1 & 1.2K & 1.2K & 4 & N/A & 8.0K & 8.9K & 6.4 & 7.2 & 1.7K & 1.6K \\
    & Test & 1.0 & 1.1K & 1.1K & 41 & N/A & 7.8K & 8.6K & 6.8 & 7.5 & 1.7K & 1.6K \\
    \midrule
    & Train & 3.0 & 2.1K & 2.1K & 4 & N/A & 23.0K & 27.2K & 11.0 & 13.0 & 8.2K & 4.4K \\
    Mn & Dev & 2.5 & 1.6K & 1.4K & 22 & N/A & 17.9K & 21.6K & 11.1 & 13.3 & 6.2K & 3.5K \\
    & Test & 2.9 & 1.8K & 1.6K & 204 & N/A & 20.2K & 24.1K & 11.0 & 13.1 & 6.8K & 3.8K \\
    \midrule
    & Train & 4.0 & 2.3K & 2.3K & 9 & 6 & 50.8K & 37.9K & 22.1 & 16.5 & 2.6K & 8.2K \\
    Zh & Dev & 3.5 & 2.0K & 2.0K & 24 & 13 & 44.0K & 33.6K & 22.5 & 17.2 & 2.6K & 7.6K \\
    & Test & 3.7 & 2.0K & 2.0K & 244 & 22 & 43.6K & 33.0K & 22.1 & 16.7 & 2.6K & 7.5K \\
    \end{tabular}
    \caption{Basic statistics of CoVoST and TT evaluation set. Token statistics are based on Moses-tokenized sentences. Speaker demographics is partially available.}
    \label{tab:covost_stats}
\end{table*}
\section{Data Collection and Processing}

\subsection{Common Voice (CoVo)}

Common Voice~\cite[CoVo]{ardila2019common} is a crowdsourcing speech recognition corpus with an open CC0 license. Contributors record voice clips by reading from a bank of donated sentences. Each voice clip was validated by at least two other users. Most of the sentences are covered by multiple speakers, with potentially different genders, age groups or accents.

Raw CoVo data contains samples that passed validation as well as those that did not. To build CoVoST, we only use the former one and reuse the official train-development-test partition of the validated data. As of January 2020, the latest CoVo 2019-06-12 release includes 29 languages. CoVoST is currently built on that release and covers the following 11 languages: French, German, Dutch, Russian, Spanish, Italian, Turkish, Persian, Swedish, Mongolian and Chinese.

Validated transcripts were sent to professional translators. Note that the translators had access to the transcripts but not the corresponding voice clips since clips would not carry additional information. Since transcripts were duplicated due to multiple speakers, we deduplicated the transcripts before sending them to translators. As a result, different voice clips of the same content (transcript) will have identical translations in CoVoST for train, development and test splits.

In order to control the quality of the professional translations, we applied various sanity checks to the translations~\cite{guzman-etal-2019-flores}. 1) For German-English, French-English and Russian-English translations, we computed sentence-level BLEU~\cite{chen-cherry-2014-systematic} with the NLTK~\cite{bird2009natural} implementation between the human translations and the automatic translations produced by a state-of-the-art system~\cite{ng-etal-2019-facebook} (the French-English system was a Transformer \emph{big}~\cite{vaswani2017attention} separately trained on WMT14). We applied this method to these three language pairs only as we are confident about the quality of the corresponding systems. Translations with a score that was too low were manually inspected and sent back to the translators when needed. 2) We manually inspected examples where the source transcript was identical to the translation. 3) We measured the perplexity of the translations using a language model trained on a large amount of clean monolingual data~\cite{ng-etal-2019-facebook}. We manually inspected examples where the translation had a high perplexity and sent them back to translators accordingly. 4) We computed the ratio of English characters in the translations. We manually inspected examples with a low ratio and sent them back to translators accordingly. 5) Finally, we used VizSeq~\cite{wang2019vizseq} to calculate similarity scores between transcripts and translations based on LASER cross-lingual sentence embeddings~\cite{artetxe2019massively}. Samples with low scores were manually inspected and sent back for translation when needed.

We also checked the overlap between train, development and test sets in terms of transcripts and voice clips (via MD5 file hashing), and confirmed they are disjoint.

\subsection{Tatoeba (TT)}
Tatoeba (TT) is a community built language learning corpus having sentences aligned across multiple languages with the corresponding speech partially available. Its sentences are on average shorter than those in CoVoST (see also Table~\ref{tab:covost_stats}) given the original purpose of language learning. Sentences in TT are licensed under CC BY 2.0 FR and part of the audio is available under various CC licenses.

We construct an evaluation set from TT (for French, German, Dutch, Russian and Spanish) as a complement to CoVoST development and test sets. We collect (speech, transcript, English translation) triplets for the 5 languages and do not include those whose speech has a broken URL or is not CC licensed. We further filter these samples by sentence lengths (minimum 4 words including punctuations) to reduce the portion of short sentences. This makes the resulting evaluation set closer to real-world scenarios and more challenging.

We run the same quality checks for TT as for CoVoST but we do not find poor quality translations according to our criteria. Finally, we report the overlap between CoVo transcripts and TT sentences in Table~\ref{tab:overlapCoVoTT}. We found a minimal overlap, which makes the TT evaluation set a suitable additional test set when training on CoVoST.

\begin{table}[h]
    \centering
    \begin{tabular}{c|ccccc}
        CoVo split & Fr & De & Nl & Ru & Es \\
        \toprule
        Train & 1.7\% & 0.2\% & 0.2\% & 0.1\% & 0.1\% \\
        Dev & 1.0\% & 0.1\% & 0.3\% & 0.0\% & 0.1\% \\
        Test & 0.9\% & 0.3\% & 0.3\% & 0.0\% & 0.4\%
    \end{tabular}
    \caption{TT-CoVo transcript overlapping rate.}
    \label{tab:overlapCoVoTT}
\end{table}

\begin{figure}[th]
    \centering
    \includegraphics[width=\linewidth]{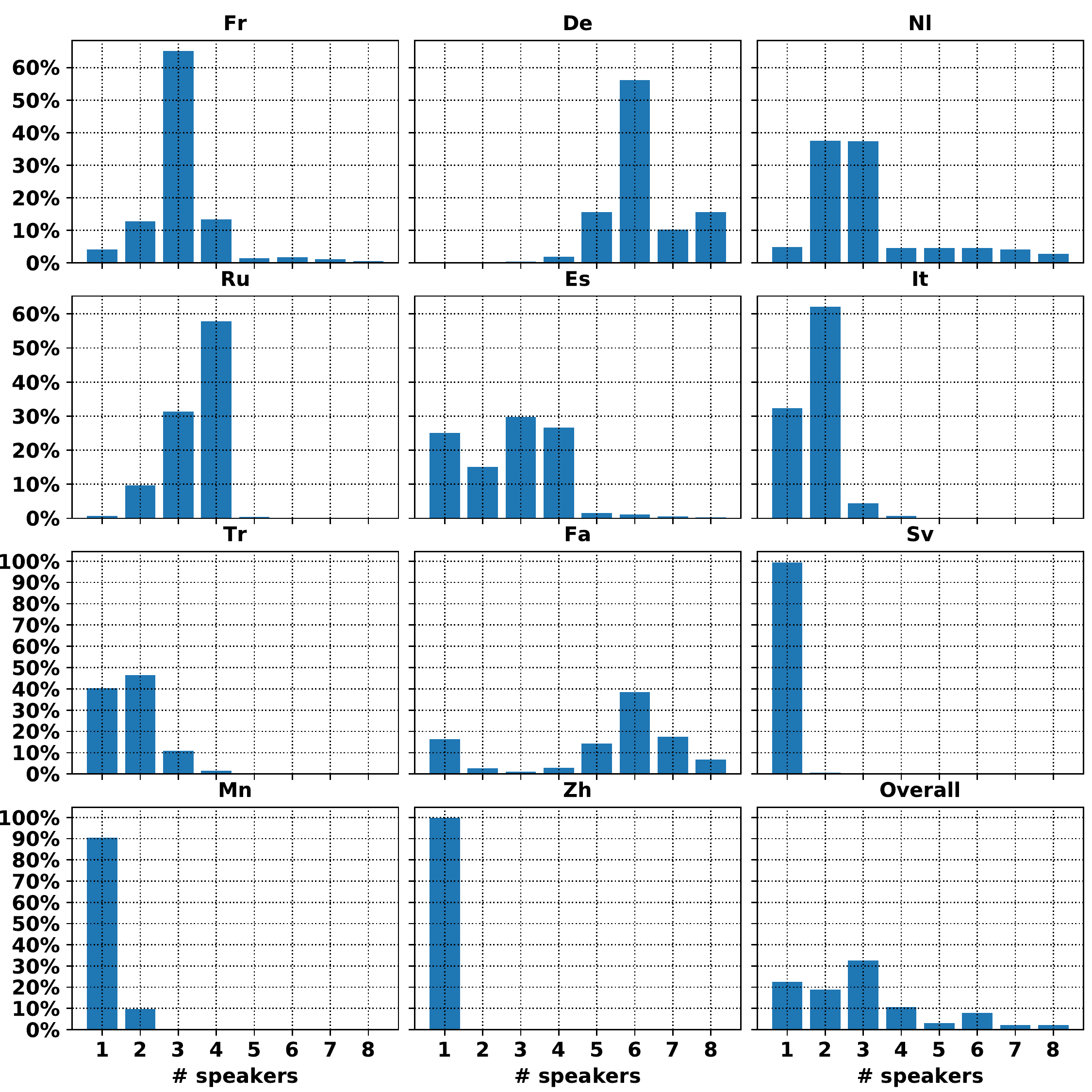}
    \caption{CoVoST transcript distribution by number of speakers.}
    \label{fig:stats_speakers}
\end{figure}

\begin{figure}[th]
    \centering
    \includegraphics[width=\linewidth]{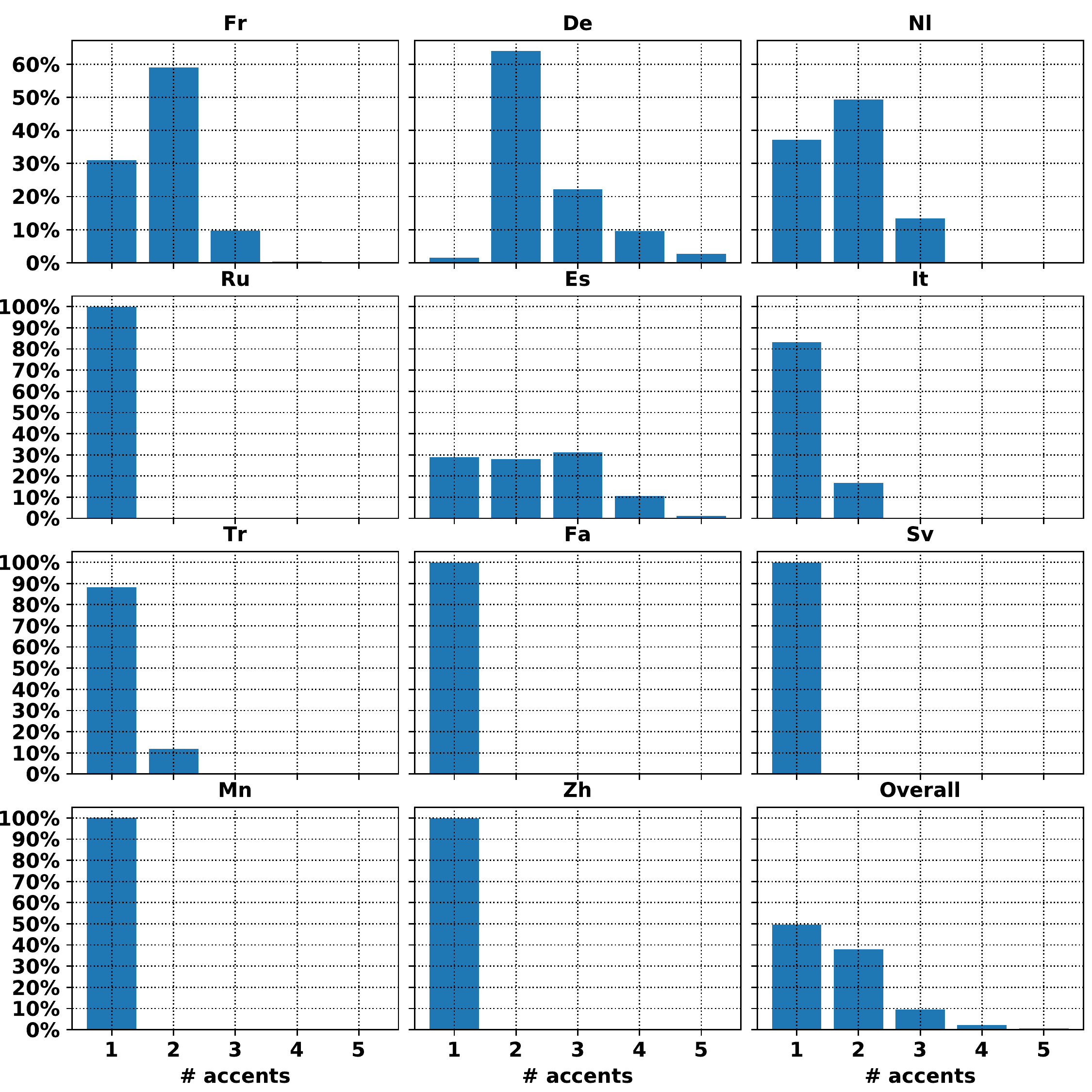}
    \caption{CoVoST transcript distribution by number of speaker accents.}
    \label{fig:stats_speaker_accents}
\end{figure}

\begin{figure}[th]
    \centering
    \small
    \includegraphics[width=\linewidth]{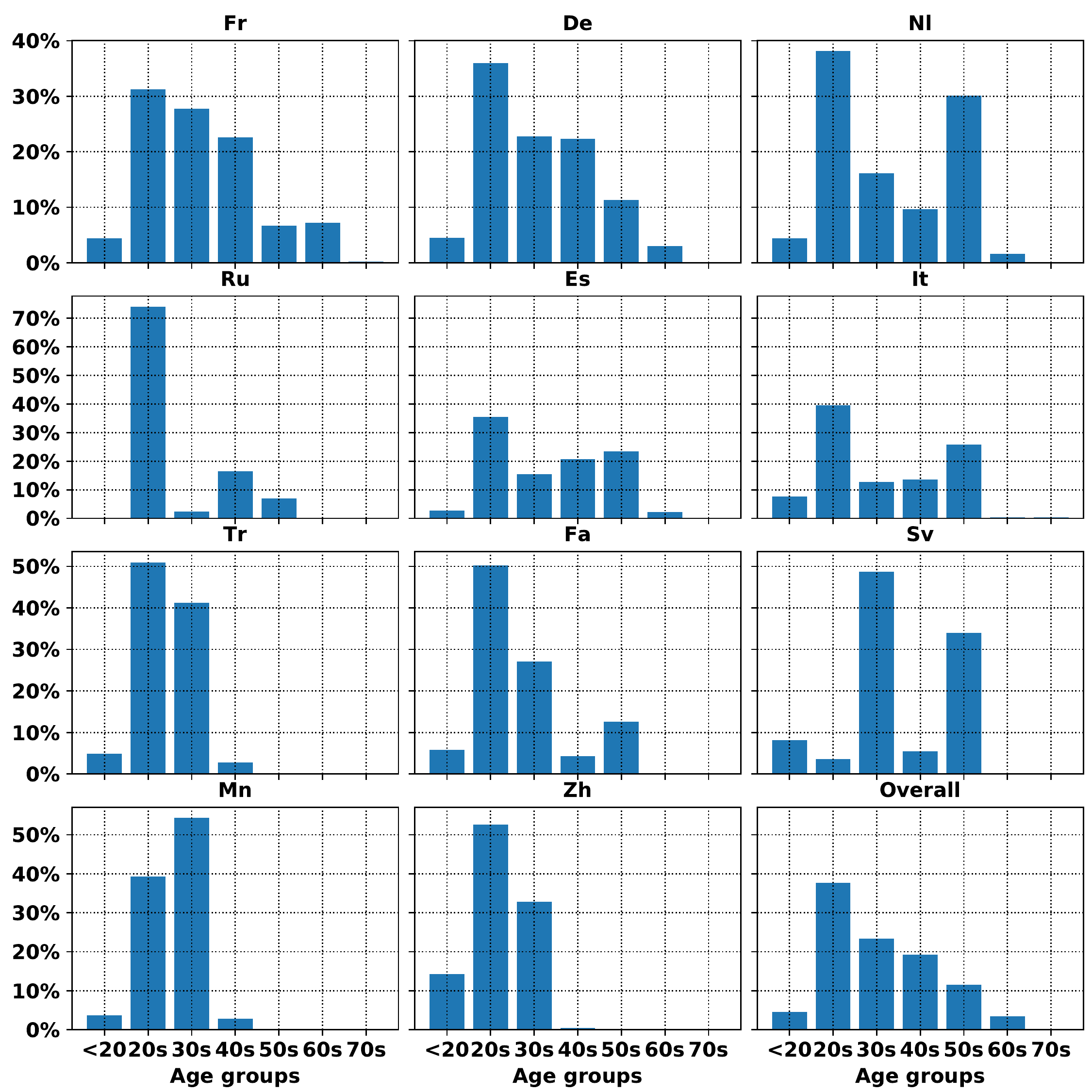}
    \caption{CoVoST transcript distribution by speaker age groups.}
    \label{fig:stats_speaker_age_groups}
\end{figure}

\section{Data Analysis}
\paragraph{Basic Statistics}
Basic statistics for CoVoST and TT are listed in Table~\ref{tab:covost_stats} including (unique) sentence counts, speech durations, speaker demographics (partially available) as well as vocabulary and token statistics (based on Moses-tokenized sentences by sacreMoses\footnote{https://github.com/alvations/sacremoses}) on both transcripts and translations. We see that CoVoST has over 327 hours of German speeches and over 171 hours of French speeches, which, to our knowledge, corresponds to the largest corpus among existing public ST corpora (the second largest is 110 hours~\cite{beilharz2019librivoxdeen} for German and 38 hours~\cite{iranzosnchez2019europarlst} for French). Moreover, CoVoST has a total of 18 hours of Dutch speeches, to our knowledge, contributing the first public Dutch ST resource. CoVoST also has around 27-hour Russian speeches, 37-hour Italian speeches and 67-hour Persian speeches, which is 1.8 times, 2.5 times and 13.3 times of the previous largest public one~\cite{cmu_wilderness}. Most of the sentences (transcripts) in CoVoST are covered by multiple speakers with potentially different accents, resulting in a rich diversity in the speeches. For example, there are over 1,000 speakers and over 10 accents in the French and German development / test sets. This enables good coverage of speech variations in both model training and evaluation.

\paragraph{Speaker Diversity}
As we can see from Table~\ref{tab:covost_stats}, CoVoST is diversified with a rich set of speakers and accents. We further inspect the speaker demographics in terms of sample distributions with respect to speaker counts, accent counts and age groups, which is shown in Figure~\ref{fig:stats_speakers}, \ref{fig:stats_speaker_accents} and~\ref{fig:stats_speaker_age_groups}. We observe that for 8 of the 11 languages, at least 60\% of the sentences (transcripts) are covered by multiple speakers. Over 80\% of the French sentences have at least 3 speakers. And for German sentences, even over 90\% of them have at least 5 speakers. Similarly, we see that a large portion of sentences are spoken in multiple accents for French, German, Dutch and Spanish. Speakers of each language also spread widely across different age groups (below 20, 20s, 30s, 40s, 50s, 60s and 70s).

\section{Baseline Results}
We provide baselines using the official train-development-test split on the following tasks: automatic speech recognition (ASR), machine translation (MT) and speech translation (ST).

\subsection{Experimental Settings}

\paragraph{Data Preprocessing} We convert raw MP3 audio files from CoVo and TT into mono-channel waveforms, and downsample them to 16,000 Hz. For transcripts and translations, we normalize the punctuation, we tokenize the text with sacreMoses and lowercase it.
For transcripts, we further remove all punctuation markers except for apostrophes. We use character vocabularies on all the tasks, with 100\% coverage of all the characters. Preliminary experimentation showed that character vocabularies provided more stable training than BPE.
For MT, the vocabulary is created jointly on both transcripts and translations. We extract 80-channel log-mel filterbank features, computed with a 25ms window size and 10ms window shift using torchaudio\footnote{https://github.com/pytorch/audio}. The features are normalized to 0 mean and 1.0 standard deviation. We remove samples having more than 3,000 frames or more than 256 characters for GPU memory efficiency (less than 25 samples are removed for all languages).

\paragraph{Model Training}
Our ASR and ST models follow the architecture in~\newcite{berard2018end}, but have 3 decoder layers like that in~\newcite{pino2019harnessing}. We pretrain their encoders on 120-hour English ASR data from Common Voice (2019-06-12 release). For MT, we use a Transformer \emph{base} architecture~\cite{vaswani2017attention}, but with 3 encoder layers, 3 decoder layers and 0.3 dropout. We use a batch size of 10,000 frames for ASR and ST, and a batch size of 4,000 tokens for MT. We train all models using Fairseq~\cite{ott2019fairseq} for up to 200,000 updates. We use SpecAugment~\cite{park2019specaugment} for ASR and ST (LB policy without time warping) to alleviate overfitting.

\paragraph{Inference and Evaluation} We use a beam size of 5 for all models. We use the best checkpoint by validation loss for MT, and average the last 5 checkpoints for ASR and ST. For MT and ST, we report case-insensitive tokenized BLEU~\cite{papineni2002bleu} using sacreBLEU~\cite{post-2018-call}. For ASR, we report word error rate (WER) and character error rate (CER) using VizSeq where both the hypothesis and reference are tokenized, lowercased and with punctuation removed.

\subsection{Automatic Speech Recognition (ASR)}

\begin{table}[t]
    \centering
    \begin{tabular}{c|cc|cc}
         & \multicolumn{2}{c|}{CoVoST Test} & \multicolumn{2}{c}{TT} \\
         & WER & CER & WER & CER \\
         \toprule
         En & 36.1 & 20.3 & & \\
         \midrule
         Fr & 24.6 & 10.7 & 43.9 & 23.7 \\
         De & 40.9 & 16.7 & 32.5 & 15.1 \\
         Nl & 56.9 & 27.2 & 53.8 & 27.0 \\
         Ru & 54.6 & 20.6 & 66.9 & 32.2 \\
         Es & 50.6 & 22.0 & 61.7 & 26.3 \\
         It & 38.2 & 14.1 & & \\
         Tr & 56.0 & 22.1 & &  \\
         Fa & 65.4 & 32.3 \\
         Sv & 82.1 & 46.5 & & \\
         Mn & 76.7 & 38.5 & & \\
         Zh & 59.2 & 33.2 & & 
    \end{tabular}
    \caption{WER and CER scores for ASR models. Non-English models are pretrained using English model's encoder.}
    \label{tab:asr_results}
\end{table}

For simplicity, we use the same model architecture for ASR and ST. Table~\ref{tab:asr_results} shows the word error rate (WER) and character error rate (CER) for ASR models. We see that French and German perform the best given they are the two highest resource languages in CoVoST. Italian is among the best as well, which is mid-resource and has limited accents. Persian is also mid-resource but is challenging because of rich speaker diversity. Most of the other languages are low resource (especially Swedish and Mongolian) and the ASR models are having difficulties to learn from this data even with pre-trained encoders.

\subsection{Machine Translation (MT)}

\begin{table}[t]
    \centering
    \begin{tabular}{c|cc}
         & CoVoST Test & TT \\
         \toprule
         Fr & 29.8 & 25.4 \\
         De & 8.0 & 8.1 \\
         Nl & 3.2 & 5.3 \\
         Ru & 3.0 & 0.7 \\
         Es & 11.0 & 2.3 \\
         It & 8.7 & \\
         Tr & 0.9 & \\
         Fa & 0.5 &  \\
         Sv & 5.0 &  \\
         Mn & 0.2 &  \\
         Zh & 5.5 & 
    \end{tabular}
    \caption{BLEU scores for MT models.}
    \label{tab:mt_results}
\end{table}

MT models take transcripts (without punctuation) as inputs and outputs translations (with punctuation). For simplicity, we do not change the text preprocessing methods for MT to correct this mismatch. Moreover, this mismatch also exists in cascading ST systems, where MT model inputs are the outputs of an ASR model. Table~\ref{tab:mt_results} shows the BLEU scores of MT models. We notice that the results are consistent with what we see from ASR models. For example thanks to abundant training data, French has a decent BLEU score of 29.8/25.4. German doesn't perform well, because of less richness of content (transcripts). The other languages are relatively low resource in CoVoST and it is difficult to train decent models without additional data or pre-training techniques.

\subsection{Speech Translation (ST)}
\begin{table}[t]
    \small
    \centering
    \begin{tabular}{c|ccccc}
         & \multicolumn{5}{c}{CoVoST Test / TT} \\
         & Fr & De & Nl & Ru & Es \\
         \toprule
         Fr & 21.4/10.9 & & & & \\
         De & & 7.6/7.5 & & & \\ 
         Nl & & & 3.4/5.0 & & \\
         Ru & & & & 4.8/1.1 & \\
         Es & & & & & 6.1/1.9 \\
         \midrule
         De+Fr & 22.1/\textbf{11.9} & \textbf{9.3}/\textbf{10.5} & & & \\
         Nl+Fr & \textbf{22.7}/\textbf{13.3} & & 2.9/\underline{3.5} & & \\
         Ru+Fr & \textbf{22.7}/\textbf{13.1} & & & \textbf{7.7}/1.0 & \\
         Es+Fr & \textbf{22.8}/\textbf{13.2} & & & & \underline{5.1}/\textbf{3.1} \\
         \midrule
         First 5 $\star$ & 21.8/11.4 & \textbf{9.8}/\textbf{12.1} & 3.4/5.7 & \textbf{7.0}/1.2 & \underline{3.9}/2.8 \\
         All 11 & 21.5/10.7 & \textbf{9.8}/\textbf{11.1} & 2.8/\textbf{6.7} & \textbf{6.2}/1.2 & \underline{4.1}/\textbf{3.4} \\
    \end{tabular}

    \caption{BLEU scores for end-to-end ST models. ST model encoders are pre-trained on English ASR. The rows indicate the languages used for training, the columns indicate the CoVoST test / TT BLEU scores on corresponding languages. Multilingual model scores that are better/worse than the bilingual baseline (by at least 1.0) are in bold/underlined. French (Fr) is highest resource among all 11 languages. $\star$ Fr, De, Nl, Ru and Es.}
    \label{tab:st_results}
\end{table}

\begin{table}[t]
    \centering
    \small
    \begin{tabular}{c|ccccccc}
         & \multicolumn{7}{c}{CoVoST Test / TT} \\
         & Fr & It & Tr & Fa & Sv & Mn & Zh \\
         \toprule
         Fr & 21.4/10.9 & & & & \\
         It & & 6.5 & & & & \\
         Tr & & & 3.1 & & & & \\
         Fa & & & & 2.8 & & & \\
         Sv & & & & & 1.9 & & \\
         Mn & & & & & & 0.3 & \\
         Zh & & & & & & & 5.6 \\
         \midrule
         It+Fr & \textbf{23.1}/\textbf{13.3} & \textbf{8.6} & & & & & \\
         Tr+Fr & \textbf{22.6}/\textbf{12.7} & & 2.4 & & & & \\
         Fa+Fr & \textbf{22.5}/\textbf{12.9} & & & 2.3 & & & \\
         Sv+Fr & \textbf{22.8}/\textbf{12.7} & & & & \underline{0.7} & & \\
         Mn+Fr & \textbf{22.8}/\textbf{13.8} & & & & & 0.3 & \\
         Zh+Fr & \textbf{22.4}/\textbf{13.2} & & & & & & \textbf{6.7} \\
         \midrule
         All 11 & 21.5/10.7 & 5.9 & \underline{1.8} & \underline{1.8} & \underline{0.9} & 0.2 & 5.1 \\
    \end{tabular}
    \caption{BLEU scores for end-to-end ST models (continuation of Table~\ref{tab:st_results}).}
    \label{tab:st_results_2}
\end{table}

CoVoST is a many-to-one multilingual ST corpus. While end-to-end one-to-many and many-to-many multilingual ST models have been explored very recently~\cite{inaguma2019multilingual,gangi2019onetomany}, many-to-one multilingual models, to our knowledge, have not. We hence use CoVoST to examine this setting.
Table~\ref{tab:st_results} and~\ref{tab:st_results_2} show the BLEU scores for both bilingual and multilingual end-to-end ST models trained on CoVoST. We observe that combining speeches from multiple languages brings gains to high-resource languages (Fr and De) consistently. Some mid-resource/low-resource languages (Ru, It and Zh) are improved as well. This includes combinations of distant languages, such as Ru+Fr and Zh+Fr. We simply provide the most basic many-to-one multilingual baselines here, and leave the full exploration of the best configurations to future work. Finally, we note that for some language pairs, absolute BLEU numbers are relatively low as we restrict model training to the supervised data. We encourage the community to improve upon those baselines, for example by leveraging semi-supervised training.

\begin{table}[t]
    \centering
    \begin{tabular}{c|cc|cc}
        & \multicolumn{2}{c|}{$\textrm{BLEU}_{MS}$} & \multicolumn{2}{c}{$\textrm{CoefVar}_{MS}$} \\
        & Breakdown & All & Breakdown & All \\
        \toprule
        Fr & & 13.38 & & 0.77 \\
        De & & 3.3 & & 2.12 \\
        Nl & & 0.81 & & 1.28 \\
        Ru & & 2.22 & & 0.67 \\
        Es & & 3.36 & & 1.02 \\
        It & & 2.46 & & 0.84 \\
        Tr & & 0.79 & & 0.80 \\
        Fa & & 1.38 & & 1.43 \\
        Sv & & 0.39 & & - \\
        Mn & & 0.03 & & - \\
        Zh & & 3.24 & & 1.0 \\
        \midrule
        De+Fr & 4.15/13.31 & 10.06 & 2.14/0.79 & 1.12 \\
        Nl+Fr & 0.70/14.01 & 12.14  & 1.54/0.80 & 0.82 \\
        Ru+Fr & 3.93/14.05 & 12.64 & 0.76/0.80 & 0.79 \\
        Es+Fr & 2.10/14.22 & 11.69 & 1.18/0.77 & 0.80 \\
        It+Fr & 3.37/14.37 & 10.99 & 0.82/0.80 & 0.80 \\
        Tr+Fr & 0.60/14.06 & 12.23 & 0.80/0.79 & 0.79 \\
        Fa+Fr & 0.82/14.07 & 11.84 & 1.50/0.78 & 0.79 \\
        Sv+Fr & 0.13/14.09 & 12.71 & -/0.80 & 0.80 \\
        Mn+Fr & 0.02/14.34 & 12.41 & 1.0/0.78 & 0.78 \\
        Zh+Fr & 4.71/14.17 & 12.65 & 0.13/0.79 &  0.79 \\
        \midrule
        First 5 & & 7.78 & & 1.19 \\
        All 11 & & 5.28 & & 1.14 \\
    \end{tabular}
    \caption{Average per-group mean and average coefficient of variation for ST sentence BLEU scores on CoVoST test set (groups correspond to one transcript and multiple speakers). The latter is unavailable for Swedish and Mongolian because models are unable to acheive non-zero scores on multi-speaker samples.}
    \label{tab:multi_speaker_variation}
\end{table}

\subsection{Multi-Speaker Evaluation}
In CoVoST, large portion of transcripts are covered by multiple speakers with different genders, accents and age groups. Besides the standard corpus-level BLEU scores, we also want to evaluate model output variance on the same content (transcript) but different speakers. We hence propose to group samples (and their sentence BLEU scores) by transcript, and then calculate average per-group mean and average coefficient of variation defined as follows:

$$
    \textrm{BLEU}_{MS}=\frac{1}{|G|}\sum_{g\in G}\textrm{Mean}(g)
$$
and
$$
\textrm{CoefVar}_{MS}=\frac{1}{|G'|}\sum_{g\in G'}\frac{\textrm{StandardDeviation}(g)}{\textrm{Mean}(g)} 
$$
where $G$ is the set of sentence BLEU scores grouped by transcript and
$G' = \{g | g\in G, |g|>1, \textrm{Mean}(g) > 0 \}$.

$\textrm{BLEU}_{MS}$ provides a normalized quality score as oppose to corpus-level BLEU or unnormalized average of sentence BLEU. And $\textrm{CoefVar}_{MS}$ is a standardized measure of model stability against different speakers (the lower the better). Table~\ref{tab:multi_speaker_variation} shows the $\textrm{BLEU}_{MS}$ and $\textrm{CoefVar}_{MS}$ of our ST models on CoVoST test set. We see that German and Persian have the worst $\textrm{CoefVar}_{MS}$ (least stable) given their rich speaker diversity in the test set and relatively small train set (see also Figure~\ref{fig:stats_speakers} and Table~\ref{tab:covost_stats}). Dutch also has poor $\textrm{CoefVar}_{MS}$ because of the lack of training data. Multilingual models may improve $\textrm{BLEU}_{MS}$ but have comparable $\textrm{CoefVar}_{MS}$.

\section{Conclusion}

We introduce a multilingual speech-to-text translation corpus, CoVoST, for 11 languages into English, diversified with over 11,000 speakers and over 60 accents. We also provide baseline results, including, to our knowledge, the first end-to-end many-to-one multilingual model for spoken language translation. CoVoST is free to use with a CC0 license, and the additional Tatoeba evaluation samples are also CC-licensed.

\section{Bibliographical References}
\label{main:ref}

\bibliographystyle{lrec}
\bibliography{lrec2020}


\end{document}